\documentclass[9pt, conference]{IEEEtran}
\renewcommand{\baselinestretch}{0.97}
\IEEEoverridecommandlockouts
\usepackage{cite}
\usepackage{amsmath,amssymb,amsfonts}
\usepackage{algorithmic}
\usepackage{graphicx}
\usepackage{textcomp}
\usepackage{xcolor}

\usepackage{caption}
\usepackage{cite}
\usepackage{amsmath,amssymb,amsfonts}
\usepackage{algorithmic}
\usepackage{graphicx}
\usepackage{textcomp}
\usepackage{xcolor}
\usepackage{multirow}
\usepackage{enumitem}
\usepackage[hyphens]{url}
\usepackage{fancyhdr}
\usepackage{colortbl}
\usepackage{hyperref}
\usepackage{hyperref}
\usepackage{wrapfig}
\usepackage{algorithm2e}
\RestyleAlgo{ruled} 
\usepackage{amssymb}
\usepackage{pifont}
\usepackage{amsmath}
\usepackage{graphicx}
\usepackage{textcomp}
\usepackage{xcolor}
\usepackage{makecell}
\usepackage{multirow}
\usepackage{pifont}
\usepackage{tikz}
\usepackage{float}
\usepackage{graphicx}
\usepackage{booktabs}
\usepackage{cleveref}
\crefformat{section}{\S#2#1#3} 
\crefformat{subsection}{\S#2#1#3}
\crefformat{subsubsection}{\S#2#1#3}
\newcommand{\cmark }{\ding{51}}%
\newcommand{\xmark }{\ding{55}}%

\def\BibTeX{{\rm B\kern-.05em{\sc i\kern-.025em b}\kern-.08em
    T\kern-.1667em\lower.7ex\hbox{E}\kern-.125emX}}

\newcommand*\circled[1]{\tikz[baseline=(char.base)]{
            \node[shape=circle,fill,inner sep=0pt] (char) {\textcolor{white}{#1}};}}
\usepackage{cite}
\usepackage{amsmath,amssymb,amsfonts}
\usepackage{algorithmic}
\usepackage{graphicx}
\usepackage{textcomp}
\usepackage{xcolor}
\def\BibTeX{{\rm B\kern-.05em{\sc i\kern-.025em b}\kern-.08em
    T\kern-.1667em\lower.7ex\hbox{E}\kern-.125emX}}
\begin{document}

\title{Accelerating LLM Inference with Flexible N:M Sparsity via A Fully Digital Compute-in-Memory Accelerator}


\author{
\IEEEauthorblockN{Akshat Ramachandran$^{*1}$, Souvik Kundu$^2$, Arnab Raha$^3$, Shamik Kundu$^3$, Deepak K. Mathaikutty$^3$, Tushar Krishna$^1$}
\IEEEauthorblockA{$^1$Georgia Institute of Technology, USA, $^2$Intel Labs, USA, $^3$Intel Corporation, USA\\
Corresponding email: \texttt{akshat.r@gatech.edu}}
\thanks{$^*$Work done during an internship at Intel.}}

\maketitle

\begin{abstract}
    Large language model (LLM) pruning with fixed N:M structured sparsity significantly limits the expressivity of the sparse model, yielding sub-optimal performance. On the contrary, support for more than one N:M pattern to provide sparse representational freedom yields a costly overhead in the hardware. To mitigate these challenges for LLMs, we first present a \underline{f}lexible \underline{l}ayer-wise \underline{o}utlier-density-\underline{a}ware N:M sparsity (FLOW) selection method. FLOW enables the identification of optimal layer-wise N and M values (from a given range) by simultaneously accounting for the \textit{presence} and \textit{distribution} of outliers, allowing a higher degree of representational freedom. To deploy the sparse models with such N:M flexibility, we then present a \underline{flex}ible low overhead, digital \underline{c}ompute-\underline{i}n-\underline{m}emory architecture (FlexCiM). FlexCiM enables support for diverse sparsity patterns by partitioning a digital CiM (DCiM) macro into smaller sub-macros which are adaptively aggregated and disaggregated through distribution and merging mechanisms for different values of N and M. Extensive experiments on both transformer-based and recurrence-based state space foundation models (SSMs) demonstrate FLOW to outperform existing alternatives with an accuracy improvement of up to $36\%$, while FlexCiM delivers up to 1.75$\times$ lower inference latency and 1.5$\times$ lower energy consumption compared to existing sparse accelerators. Code is available at: \url{https://github.com/FLOW-open-project/FLOW}
\end{abstract}


\section{Introduction}


To reduce the colossal size of large language models (LLMs) and enable their efficient deployment on resource-constrained devices, post-training pruning has emerged as an effective model compression method \cite{yin2023junk, frantar2023sparsegpt, sun2023simple}. It reduces the memory footprint of the pre-trained LLMs by removing ineffectual model parameters, at the granularity of individual weights (\textit{unstructured}) or blocks of weights (\textit{structured}), and storing sparse tensors in a compressed format (CSR/CSC) \cite{jeong2023vegeta}. 
Notably, model pruning may yield compute acceleration via skipping ineffectual computations associated with the zero-valued weight/activation. However, traditional weight pruning often requires fine-tuning, which becomes exceedingly compute-heavy for LLMs. Furthermore, this often requires the model to yield structured pruned weights, which can cause a high accuracy drop compared to the models pruned via an unstructured approach. To mitigate the dilemma of accuracy vs. compute efficiency, \textit{structured N:M sparsity} \cite{yin2023outlier, jang2021sparsity} has emerged as a popular solution that can yield models with compute acceleration benefits while maintaining good accuracy. In specific, many of the existing commodity accelerators have enabled sparse N:M acceleration for fixed N:M values \cite{mishra2021accelerating}. Therefore, this work focuses on N:M structured sparsity for efficient LLM inference.

To enable post-training pruning and mitigate fine-tuning cost, SparseGPT \cite{frantar2023sparsegpt} proposed to employ row-Hessian information to rank the importance of pre-trained weights and prune the least important ones.  SparseGPT still required high compute owing to its iterative computation of second-order information (Hessian) for each layer. Wanda \cite{sun2023simple} avoided the computationally expensive second-order details by introducing a pruning metric that considers both the weight magnitudes and their corresponding input activations. Considering the performance benefits of structured N:M sparsity, both these works \cite{frantar2023sparsegpt, sun2023simple} extended their approach to support a fixed and pre-determined N:M pattern for all the layers.  However, only recently \cite{yin2023outlier} researchers have determined that due to the presence of a large number of outliers\footnote{Scalars with larger magnitude compared to the rest of the tensor.} in LLMs, different layers are inherently heterogeneous and exhibit different tolerances to pruning.  Based on this, {outlier weighed layer-wise sparsity (OWL)} \cite{yin2023outlier} was proposed. OWL differs from prior techniques in its assignment of different sparsity ratios for different layers through allocations of different N values in the N:M pattern with fixed M. However, this is still sub-optimal and has limited representational freedom due to their static selection of M value across all the layers. Moreover, existing hardware accelerators cost significant overhead to employ different values of N and/or M patterns in a single architecture, which was beyond the scope of \cite{yin2023outlier} to address.

 It is well established that the decode stage of LLM inference is memory bound \cite{ramachandran2024microscopiq, agrawal2024metron}, with weights loading constituting a significant bottleneck. Traditional Von-Neumann accelerators consume significant energy and time when moving weights from on-chip buffers to processing elements (PEs). This makes existing traditional accelerators with flexible structured sparsity support \cite{jeong2023vegeta, wu2023highlight} poor candidates to accelerate N:M sparse LLMs for inference. An alternative computing paradigm, namely, \textit{digital compute-in-memory} (DCiM) offers an efficient solution to the above challenge by moving the compute within the memory arrays. This mitigates the memory access bottleneck. However, despite the advantages of DCiM over traditional digital accelerators, there has been limited exploration of flexible structured sparsity support for DCiM-based architectures. This can be associated with the rigid crossbar structure of memory arrays and the infeasibility of transferring the flexible structured sparsity techniques employed for digital accelerators to DCiM (\cref{sec:motivation}).

\begin{figure}[t]
    \centering
    \includegraphics[width=0.9\columnwidth, keepaspectratio]{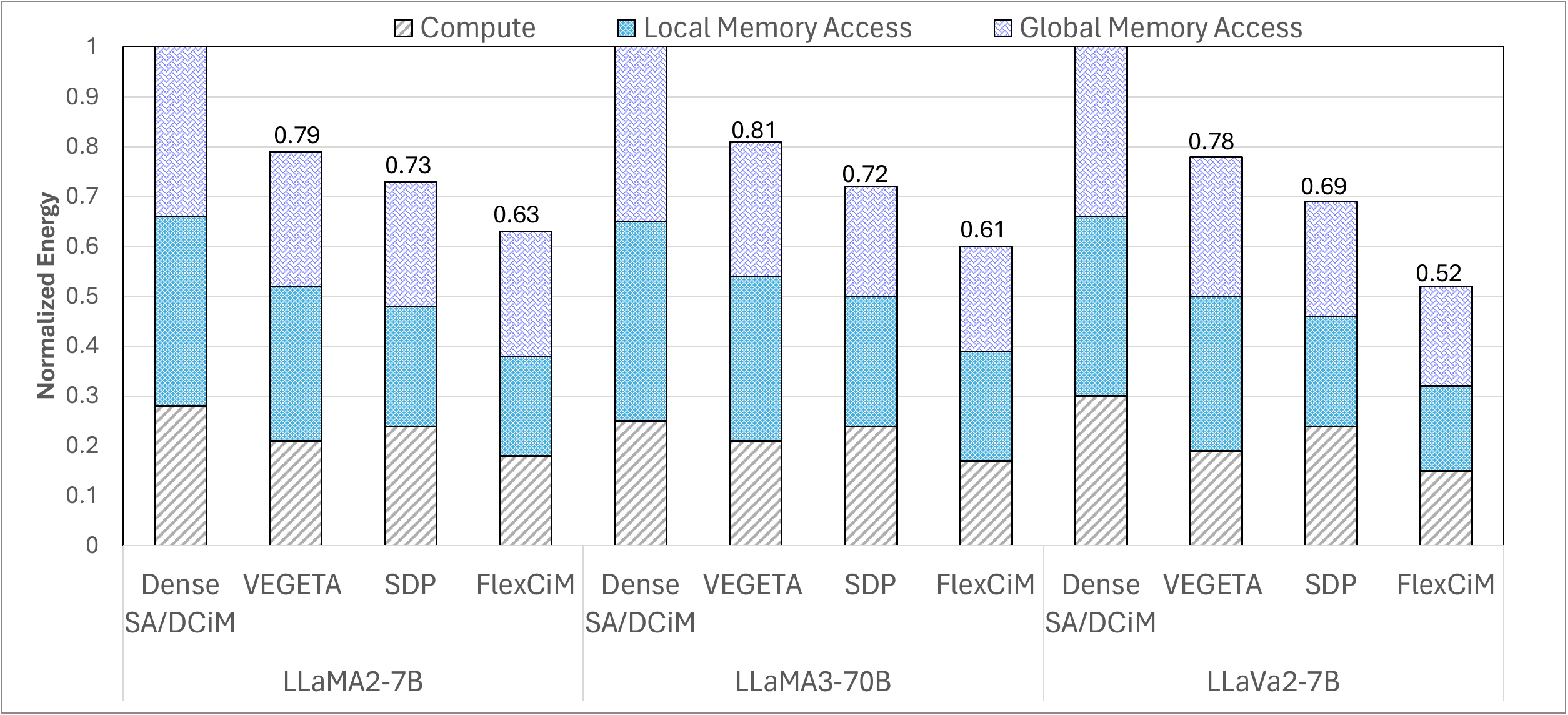}
        \vspace{-1mm}
        \caption{Normalized energy comparison between several digital and DCiM accelerators and the proposed FlexCiM across different models for flexible N:M inference acceleration.}
        \label{fig:energy_CiM}
        \vspace{-6mm}
\end{figure}

\noindent
\textbf{Our contributions. } We present two significant contributions toward addressing the above-mentioned algorithmic and hardware limitations. Specifically, we present a pruning algorithm that can assign different N:M sparsity with a higher degree of freedom and then present the first DCiM accelerator to support such patterns with minimal overhead, as summarized below.  

We first explore the efficient and optimal assignment of the N:M sparsity pattern to each layer of an LLM. Specifically, we identify that compared to static N or M, \textit{having the luxury to choose both N and M allows us to yield better sparse representational freedom necessary to maintain improved model accuracy}. To this end, we propose FLOW--\underline{f}lexible \underline{l}ayer-wise \underline{o}utlier-density-\underline{a}ware N:M sparsity selection method. FLOW is based on a key insight that \textbf{both the presence of outliers and their distribution in different layers} contributes to the layer heterogeneity, impacting their tolerance towards model pruning. We leverage these two characteristics--presence and distribution of outliers to assign diverse N:M values with the ability to determine the optimal N and M for a layer simultaneously. 

Towards deploying the flexible N:M sparse LLMs in DCiM, we first analyze its challenges in directly incorporating techniques employed by digital accelerators such as \cite{jeong2023vegeta, wu2023highlight}, particularly the infeasibility of incorporating large multiplexers within each memory cell of DCiM due to the rigid crossbar structure of the memory array. We then introduce a novel DCiM-based accelerator with flexible structured sparsity support--FlexCiM. It decomposes an existing DCiM macro into multiple partitions called sub-macros (partitioned along the row dimension) and introduces two new units--distribution and merging units. These units coordinate the mapping of weights to each sub-macro and broadcasting of input activations (\textit{iAct}s) to support diverse N:M sparsity with minimal overheads. 

We conduct extensive experiments with foundation models based on \textbf{both transformer and state space model (SSM) primitives}. Specifically, experiments across various LLMs and VLMs demonstrate FLOW outperforming SoTA pruning techniques by achieving up to \textbf{36}\% better accuracy at high sparsity.  Our accelerator, FlexCiM, on the other hand, can yield up to \textbf{1.75}$\times$ lower inference latency with \textbf{1.5}$\times$ lower energy (see \autoref{fig:energy_CiM}) compared to the SoTA sparse accelerators \textbf{with a minimal area overhead of $\sim 6\%$}.


\section{Background and Related Works}
\noindent
\textbf{LLM pruning with N:M Sparsity. } 
N:M sparsity retains N non-zero elements within a fine-grained block of consecutive M elements. Existing LLM pruning works, including SparseGPT \cite{frantar2023sparsegpt} and Wanda \cite{sun2023simple}, when extended to structured N:M sparsity, were limited in their exploration to a fixed N:M pattern for all layers. More recently, \cite{yin2023outlier} highlighted the importance of a heterogeneous sparsity budget for different LLM layers. Specifically, it proposed a non-uniform layer-wise N:M sparsity scheme with different N values for different layers determined based on the number of outliers in each layer. However, unlike existing methods \cite{frantar2023sparsegpt, sun2023simple, zhang2023dynamic, yin2023outlier, yin2023junk}, in this work, to yield more optimally pruned models, we present a higher degree of sparse representational freedom by identifying both the optimal N and M values of an N:M pattern.      


\noindent
\textbf{DCiM accelerators. }Conventional digital accelerators \cite{ramachandran2024algorithm, ramachandran2024microscopiq, parashar2017scnn} employ compute engines separate from memory (buffers) to handle LLM inference, often leading to memory access bottlenecks, increasing the memory access cost \cite{sumbul2023fully}. Furthermore, the decode stage of LLM inference is primarily memory-bounded, making weight/activation loading a significant bottleneck \cite{lin2024awq}. In light of these, CiM architectures \cite{yu2021compute, sumbul2023fully, fujiwara20225} have been propelled into the mainstream. In CiM, the processing elements and weight local memory (SRAM, ReRAM, DRAM, etc.) are merged into a single macro \cite{tu2022sdp, kim2021colonnade}, effectively reducing on-chip memory access, yielding accelerated computation. Early work on CiM focused on analog CiM architectures \cite{liu202033,chen202115}, embedding analog multiply-accumulate (MAC) units within memory arrays. While analog CiM architectures demonstrate high energy efficiency, they have various limitations, including vulnerability to noise/process variations, limited bit precision, and significant ADC/DAC overheads, collectively restricting their scalable adoption. Recently, DCiM architectures have gained considerable traction, with various industry accelerators \cite{tu2022sdp, sumbul2023fully, fujiwara20225} adopting these macros to embed digital MAC within memory directly. This eliminates the need for costly DACs/ADCs and offers enhanced robustness against variation and noise and sufficient precision to achieve high efficiency and accurate inference. Additionally, DCiM architectures are inherently compatible with existing digital design flows, making them suitable candidates for LLM acceleration.

\noindent
\textbf{Accelerators supporting sparsity. }Among digital accelerators, the NVIDIA sparse tensor core \cite{mishra2021accelerating} exploited fixed 2:4 sparsity of weights in DNNs. VEGETA \cite{jeong2023vegeta} presented flexible N:M sparsity support for weights, and S2TA \cite{liu2022s2ta} extended support for the same for both weights and activations, albeit with significant overheads. \cite{wu2023highlight} proposed hierarchically structured sparsity to represent diverse sparsity patterns in DNNs. Flexible architectures like \cite{raha2024flexnn} can potentially accelerate any N:M ratio exploiting unstructured sparsity, yet often suffer from unbalanced critical path delay. However, support for flexible structured sparsity in DCiM accelerators has been largely unexplored. The majority of sparse DCiM accelerators have either focused on bit-level sparsity \cite{zhan202428, duan2024towards} or unstructured sparsity \cite{zhong2023digital, sridharan2024sp}. Apart from these, a few works \cite{tu2022sdp, yue202115} introduced fixed 1:2 sparsity support in DCiMs. However, challenges and potential for deploying flexible N:M sparsity in DCiM are yet to be unveiled.

\section{Motivation}
\label{sec:motivation}
\begin{figure}[t]
    \centering
    \includegraphics[width=\columnwidth, keepaspectratio]{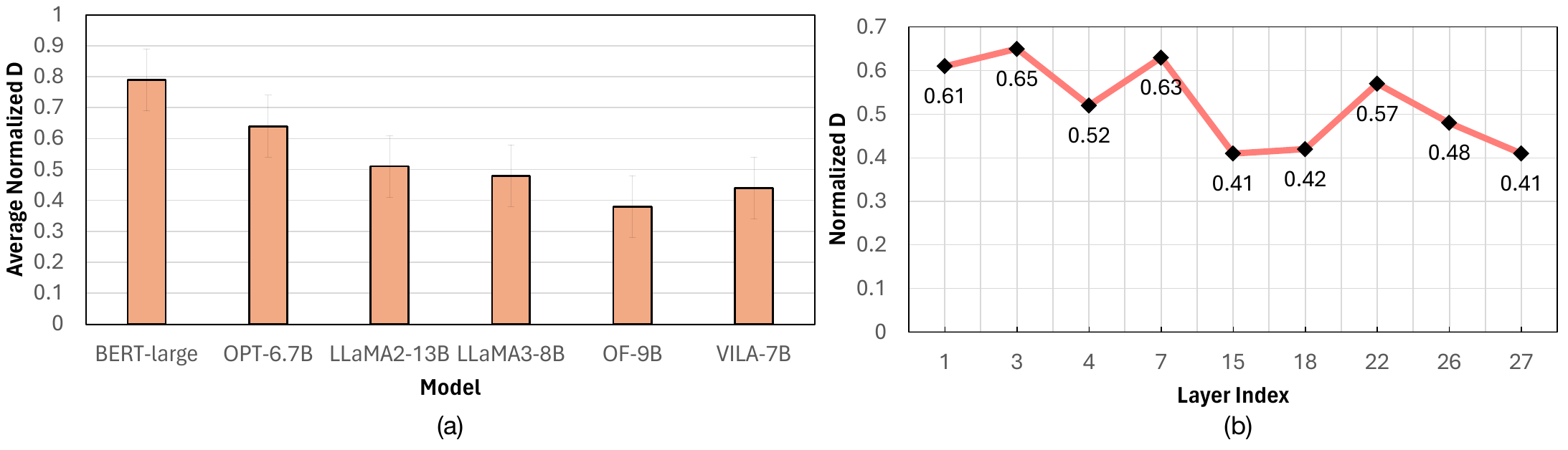}
        \vspace{-6mm}
        \caption{(a) Outlier distribution measured by pairwise ${L}_1$ distance between outliers (normalized to BERT-large) across different models. (b) Intra-model variations in outlier distribution across different layers of a LLaMA3-8B model.}
        \label{fig:motivation}
        \vspace{-7mm}
\end{figure}

\subsection{Distribution of Outliers in LLMs}
\label{sec:motivation_outlier}
OWL \cite{yin2023outlier} presented a non-uniform layer-wise N:M sparsity with a fixed M and a different N for different layers with N being proportional to the layer outlier ratio. That is, layers with a high outlier ratio would have a high N value. However, varying N with a fixed M value, results in suboptimal sparse representation freedom in each layer. We argue that a predetermined M value when assigning different N:M sparsity patterns ignores another key metric in LLM weights, namely the \textbf{outlier distribution} ($D$). For a tensor $l$, we compute $D^l$ by a \textit{proxy metric that measures the summation of the pairwise distance between the outliers averaged over the number of outliers}. In Eq. \ref{eq:out_dist}, \textit{dist}(.) measures the pair-wise $L_1$-distance between two outliers, where $^nC_2$ is the total number of outlier pairs in a layer with $n$ outliers. We then compute the \textbf{normalized outlier distribution} ($ND$) through the min-max normalization across all the layers as shown in Eq. \ref{eq:out_dist}.
\begin{align}
    ND^l = \frac{(D^l - min(\mathcal{D}))}{max(\mathcal{D})}, \text{where} \quad D^l = \frac{\sum_1^{^nC_2}{\textit{dist}{(o_k, o_j)}}}{^nC_2}
    \label{eq:out_dist}
\end{align}
where $\mathcal{D}$ represents the set of Ds for all the layers.
To understand the variance of $ND$ over different models, in \autoref{fig:motivation}(a) we plot the average $ND$ averaged over all the layers for different models (including VLMs).  In specific, we observe that models like BERT, OPT have a relatively high average $ND$, indicating that outliers are sparsely distributed within these models. On the contrary, models like LLaMA3-8B and OpenFlamingo-9B possess a significantly low value of average $ND$, implying that outliers are largely clustered together. 
Furthermore, even within the same model, different layers depict varied normalized outlier distributions. \autoref{fig:motivation}(b) demonstrates the diverse $ND$ values for different layers of a LLaMA3-8B. We hypothesize that while choosing the N:M sparsity patterns to different layers, it is important to choose the optimal M value for each layer. If a layer has largely clustered outliers (for example: layer $\#$ 15 of LLaMA3-8B), the pruning algorithm should favor a larger M value when assigning an N:M sparsity pattern. This will help to (a) minimize the possibility of pruning outliers within a block of M and (b) have a higher degree of flexibility in deciding which features to prune. We demonstrate the usefulness of such N:M selection in yielding a better perplexity score in \autoref{tab:llm_results} compared to fixed M.

\subsection{Challenges of Flexible Structured Sparsity in DCiMs}
\label{sec:cim_motivation}
Sparse digital accelerators \cite{jeong2023vegeta, mishra2021accelerating} have significant multiplexer overhead within each PE to support flexible N:M sparsity. However, this strategy is not directly applicable to DCiM-based accelerators due to various reasons:

\noindent
\textbf{Area overhead.} The key benefit of DCiM architectures is their high memory density \cite{sumbul2023fully, tu2022sdp}. Typical memory cells in DCiM employ 6T or 12T SRAM cells \cite{fujiwara20225}. However, a single 8:1 multiplexer for flexible N:8 support requires over 42 transistors \cite{tu2022sdp} \textit{increasing the area overhead by more than $3\times$} and undermining the compactness and efficiency of the DCiM architectures.

\noindent
\textbf{Rigid crossbar structure. }Digital accelerators using weight-stationary dataflow stream up to M activations into each PE \cite{jeong2023vegeta}, enabling PE-level selection. In contrast, DCiM architectures are restricted by the two-bit line (BL and $\bar{\text{BL}}$) structure of SRAM cells, allowing only two activations to be streamed into a DCiM macro.

\noindent
\textbf{Bit-serial processing. }DCiM architectures employ bit-serial operations to maximize energy efficiency, where each SRAM bit-cell stores a single weight bit and computes partial products with serially streamed activation bits. Von-Neumann accelerators, by contrast, use parallel multipliers/adders and store metadata alongside weights in each PE to manage activation selection. For N:8 sparsity, each N non-zero element requires 3 bits of metadata. In DCiMs, \textit{this would necessitate more metadata bits per SRAM cell than weight bits, complicating implementation}.
These challenges highlight the need for a novel approach to enable flexible N:M sparsity tailored for DCiMs. 

\section{FLOW N:M sparsity}
\label{sec:flow}

\begin{figure}[t]
    \centering
    \includegraphics[width=\columnwidth, keepaspectratio]{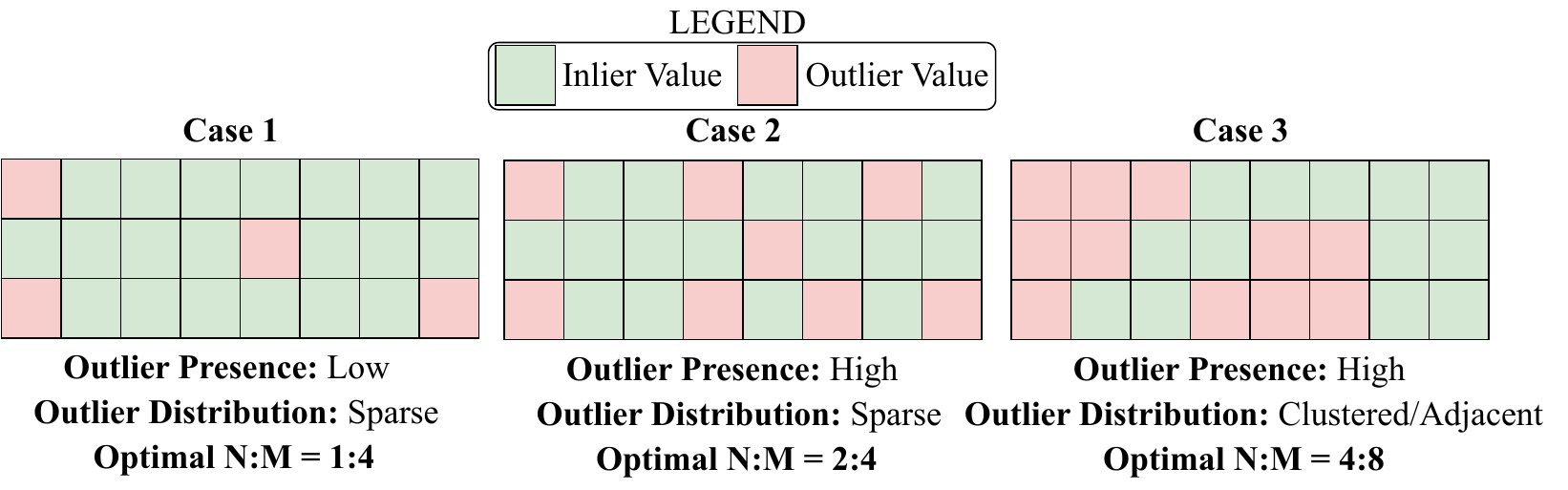}
        \vspace{-6mm}
        \caption{Example of efficient N:M assignment based on outlier presence and distribution in different situations.}
        \label{fig:algorithm_flow}
        \vspace{-6mm}
\end{figure}
In this section, we first formalize outlier identification and then describe FLOW's methodology for optimal N:M assignment.  

\subsection{Outlier Identification}
\label{sec:outlier_identification}
In this work, we target N:M sparsity for weights \cite{yin2023outlier} and focus our discussion on weight outliers. Let us assume an LLM weight tensor $W \in \mathbb{R}^{(C_{in} \times C_{out})}$ with input activations $X \in \mathbb{R}^{(T*K \times C_{in})}$. Here, $T, K$ are the number of tokens and sequence length, respectively. Inspired by prior works \cite{sun2023simple}, we assign an importance score to each weight element $W_{ij}$, determined as $\mathcal{I}_{W} = |W_{ij}|\cdot ||X_j||_2$ with $||X_j||_2$ being the  ${L}_2$ norm of the input activations connected to the weight element.  $\mathcal{I}_{W}$ acts as a measure to identify the outliers based on both weight and associated activation values \cite{sun2023simple}. Upon obtaining this score, we calculate the \textit{mean} ($\mu$) and \textit{standard deviation} ($\sigma$) of the weight importance scores for each layer \cite{yin2023outlier}. Subsequently, we categorize all weights exceeding the threshold $\tau \cdot \sigma$ as outliers and the rest as inliers. We empirically identify $\tau=3$ or $\tau=5$ provide the best results.

\subsection{FLOW Methodology}
For an $L$-layer LLM, with a target sparsity ratio of $\mathcal{S}$, the layer-wise sparsity ratios are assigned and handled by two lists $\mathcal{N} = [N^1, N^2, ..., N^L]$ and $\mathcal{M} = [M^1, M^2, ..., M^L]$. For a layer $l$, the sparsity pattern is $N^l$:$M^l$, the sparsity ratio is $S^l = (M^l - N^l)/(M^l)$ and $\mathcal{S} = (\sum_{l=1}^{l=L}S^l)/L$. Following the process outlined in \cref{sec:outlier_identification}, we obtain the outlier fraction in each layer to get $\mathcal{O} = [O^1, O^2, ..., O^L]$, where $O^l$ identifies the normalized outlier fraction in a layer $l$ \cite{yin2023outlier}. Intuitively, if a layer has a larger $O^l$, the number of non-zeros in a block of $M^l$ elements should be higher (i.e., $N^l\uparrow$). In specific, $N^l \propto O^l$. In \autoref{fig:algorithm_flow}, examples with higher presence of outliers have larger values of $N$. 

FLOW takes the normalized outlier distribution of each layer ($ND^l$) into account to identify the optimal $M^l$. In specific, for each layer weight tensor, we partition it into $B$ non-overlapping blocks of dimension $(128\times128)$. Within a block $b$, we measure $D^l_b$ by calculating the pairwise ${L}_1$ distance between all the outlier pairs following \autoref{eq:out_dist}. We obtain $D^l$ as $\sum_{b=1}^B D^l_b$, and finally get $ND^l$ via normalizing the $D^l$ values following \autoref{eq:out_dist}. A higher $ND^l$ indicates a farther distance between outliers within a layer, implying that they are sparsely distributed (Case 1 in \autoref{fig:algorithm_flow}), whereas smaller $ND^l$ indicates a large number of closely located outliers (Case 3 in \autoref{fig:algorithm_flow}). Intuitively, if a layer has many closely located outliers, a higher value of $M^l$ is preferred, providing more freedom while performing weight pruning and preventing inadvertent pruning of outliers. In specific, $M^l \propto (1-ND^l)$.

\noindent
\textbf{Automated layer-wise N, M allocation.} Given the relations of $N^l, M^l$, we formulate the problem of assigning layer-wise N:M values as an \textit{integer linear programming} (ILP) problem:
\vspace{-2mm}
\begin{gather}
\label{eq:sparsity_constr}
\text{argmin} \sum_{l=1}^{L} \left( \alpha \cdot |N^l - k \cdot O^l| + \beta \cdot |M^l - h \cdot (1 - ND^l)| \right) \\ \nonumber
s.t. \quad \frac{1}{L} \sum_{l=1}^{L} S^l = S, 0 \leq S^l \leq 1, \forall l \in \{1, \dots, L\} \\ \nonumber
1 \leq N^l \leq M^l, \quad N^l \in \{1, 2, 4, 8 \}, \quad M^l \in \{2, 4, 8 \}
\end{gather}

where $\alpha, \beta$ are the weighing factors to optimize N and M values, respectively. $k$ and $h$ are the hyper-parameters to keep the terms $O^l$ and $ND^l$ in the same ballpark range as $N^l$ and $M^l$ choices, respectively. We empirically choose their values to be $(\alpha, \beta,k, h) = (1,4, 8, 8)$. This formulation aims to assign values to $N^l$ and $M^l$ for each layer while minimizing deviations from desired characteristics, all while maintaining the target sparsity (\autoref{eq:sparsity_constr}). $N^l$ is aligned with the normalized outlier count $O^l$, and $M^l$ is aligned with the outlier distribution $(1-ND^l)$. It is important to note that FLOW can be employed to explore any combination of N and M. However, to ensure hardware efficiency, we restrict N and M choices to powers of two and a maximum value of 8 for this work (see N, M choice set mentioned in \autoref{eq:sparsity_constr}). 

\begin{figure}[t]
    \centering
    \includegraphics[width=0.95\columnwidth, keepaspectratio]{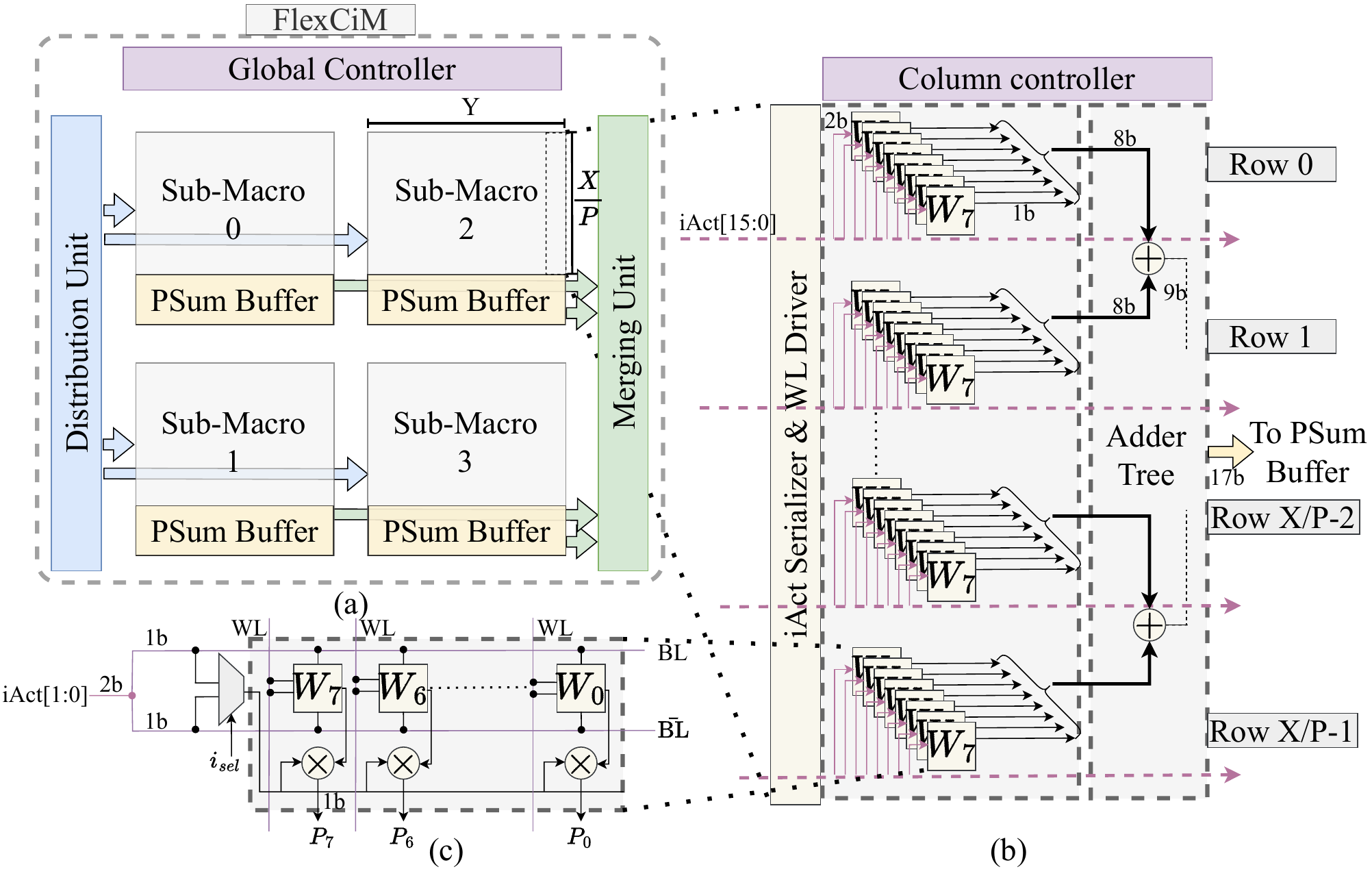}
        \vspace{-3mm}
        \caption{(a) FlexCiM overview with a partition size $P=4$; (b) Organization of a single column of a FlexCiM sub-macro; (c) Memory cell structure.
        }
        \label{fig:flexcim_overview}
        \vspace{-7mm}
\end{figure}

\section{FlexCiM Accelerator}
\noindent
\textbf{Overview. }\autoref{fig:flexcim_overview}(a) provides an overview of the FlexCiM architecture. FlexCiM extends an existing $X \times Y \times 8$ \cite{tu2022sdp} SRAM-based DCiM bit-cell macro supporting fixed 1:2 sparsity, where $(X, Y)$ represent the crossbar array dimensions with the memory word size being $8$-bits. FlexCiM can be employed to accelerate any layer that can be represented as a GEMM \cite{jeong2023vegeta} operation. It partitions the existing macro into $P$  components (named as sub-macros), each of dimension $\frac{X}{P} \times Y \times 8$. To seamlessly support diverse N:M sparsity patterns with this architecture, we introduce two novel hardware components, namely the \textit{distribution} and the \textit{merging} units to orchestrate the working of the $P$ sub-macros. To enable the bit-serial computation, a local input activation (\textit{iAct}) buffer feeds the activations to the distribution unit (detailed later) at a rate of $X$ 8-bit activations per cycle, which is serialized via the \textit{iAct} serializer unit in each sub-macro \cite{sumbul2023fully, fujiwara20225}. We employ a two-tier control unit comprising a global controller and column-wise controller to generate control signals that manage pipelining across rows and columns of sub-macros and configure multiplexer select signals. In our demonstrated implementation, we assume $X = 128$,$ Y =32$ for the DCiM bit-cell macro, with partition size of $P=4$, resulting in an 8Kb sub-macro of dimensions $32\times 32 \times 8$. Each sub-macro column is divided into $32$ banks to enable parallel MAC operation. 

\subsection{Sub-Macro Architecture}
\noindent
\textbf{Memory cell. } \autoref{fig:flexcim_overview}(c), presents a memory word in FlexCiM, with each word composed of 8 bit-cells. \emph{To abstract the circuit complexity and demonstrate our proposal to be orthogonal to any SRAM technology, we follow \cite{li2023laxor} and implement a 28nm, standard cell, latch-based memory structurally resembling a 6T SRAM cell \cite{tu2022sdp}.} Each memory cell has two bit-lines $\text{BL}, \bar{\text{BL}}$, shared by all cells in a row, and a word-line (WL), shared per column. All memory cells in a column share a control signal to enable the compute units, namely \textit{EN\_COL}. 
Inspired by \cite{tu2022sdp}, we implement a 2:1 multiplexer that is shared by a memory word to select one of the two \textit{iActs} (via control signal ($i_{sel}$) streamed along the two bit-lines to enable 1:2 structured sparsity as a baseline case. A bit-serial multiplier within each SRAM cell (implemented as NOR gate) performs the multiplication between the selected \textit{iAct} and stored weight. The memory cell has two modes of operation: memory ($\mathcal{M_S}$) and compute ($\mathcal{C}$) mode. Standard read and write operations are performed in $\mathcal{M_S}$ mode. The WL is activated to access all cells in a column, with BLs used to read/write data. $\mathcal{C}$ mode is enabled for all cells in a column only when WL=0 and the corresponding \textit{EN\_COL}=1, allowing the compute units to perform MAC operations.

\noindent
\textbf{Memory column. }\autoref{fig:flexcim_overview}(b) illustrates the organization of a column in a sub-macro. Each column comprises 32 rows of 8-bit memory words and a column-wise adder tree. Each memory word performs bit-wise multiplication with the streamed \textit{iAct} bits (MSB first), which are then accumulated column-wise by a 32-input adder tree and directed to the partial sum (PSum) accumulator.

\noindent
\textbf{Partial sum accumulator. }Since DCiMs typically perform bit-serial arithmetic, the MACs from each column-wise adder tree must be bit shifted based on the bit significance of the streamed \textit{iAct} and subsequently accumulated to compute the final partial sum. The complete operation can be expressed as, $\text{PSum}_{\text{final}} = \sum_{i=0}^{i=7} 2^i \times \sum(\textit{iAct}[i] \cdot W)$.

\noindent
\textbf{Column-wise controller. }Each sub-macro possesses a controller that is dedicated to generating the $i_{sel}$ signal for all the 2:1 multiplexers in each column and enabling column pipelining (see \cref{sec:pipeling}) via the \textit{EN\_COL} signal.

\subsection{Hardware Extensions for Flexible N:M Sparsity}
\begin{figure}[t]
    \centering
    \includegraphics[width=0.85\columnwidth, keepaspectratio]{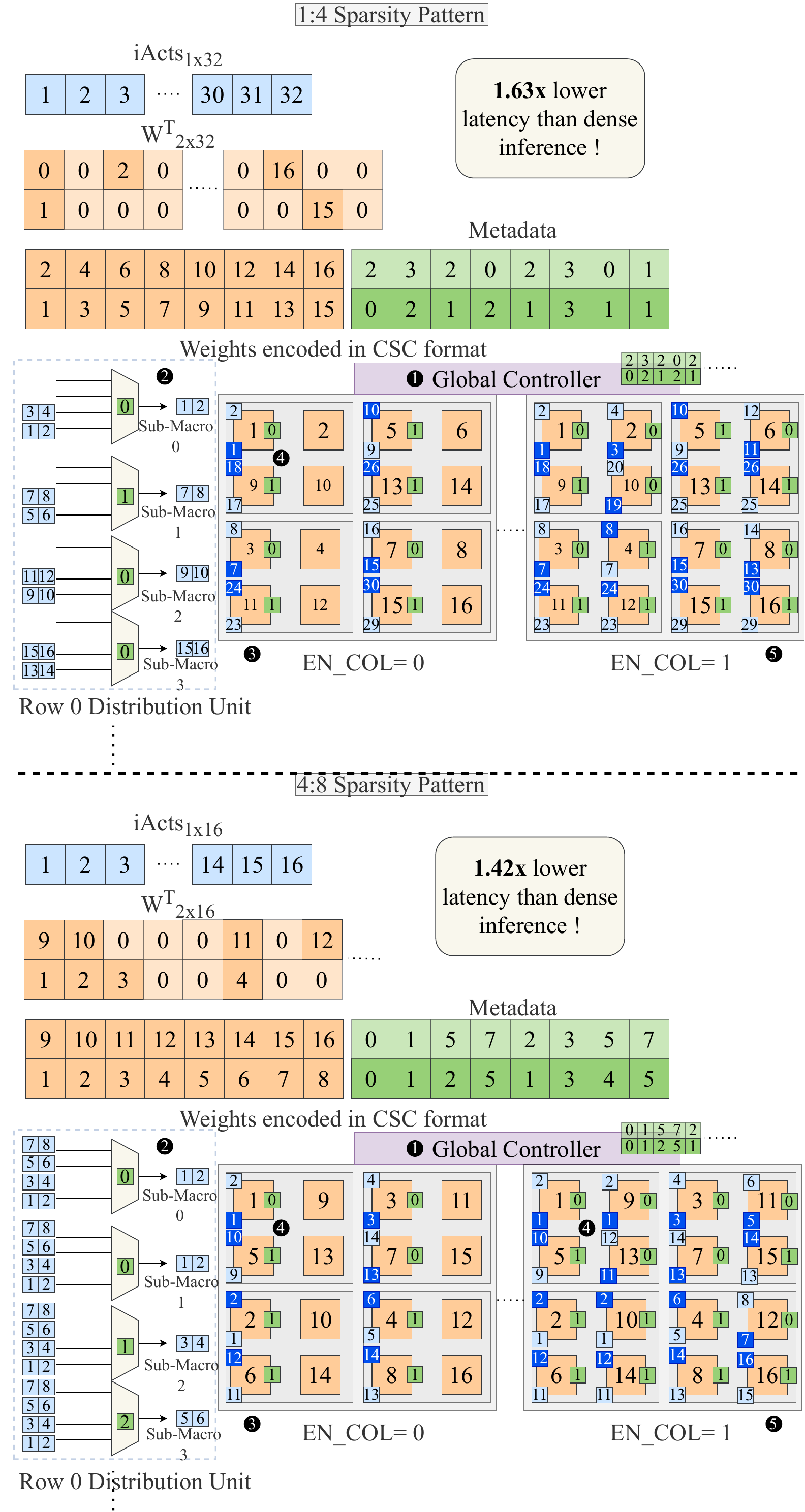}
        \vspace{-3mm}
        \caption{Example of FlexCiM running 1:4 and 4:8 sparsity patterns.}
        \label{fig:flexcim_example}
        \vspace{-6.5mm}
\end{figure}

\noindent
\textbf{N:M sparsity storage format. }In FlexCiM all sparse weight tensors are encoded in compressed sparse column (CSC) format \cite{jeong2023vegeta}. Each column is stored as a list consisting of all the nonzero values, and the coordinates of the N non-zero values within a block of size M are stored as the corresponding metadata (see \autoref{fig:flexcim_example}).

\noindent
We introduce two key hardware blocks, the distribution unit and merging unit, that enable flexible N:M structured sparsity together.

\noindent
\textbf{Distribution unit. }The distribution unit is responsible for efficiently feeding \textit{iActs} to the sub-macros based on the N and M values. As depicted in \autoref{fig:flexcim_overview}, it is composed of $P$ $P:1$ multiplexers that serve each of the $P$ sub-macros ($P=4$ in our implementation) with a bit-width of 16 bits (packing two 8-bit \textit{iAct}s) per input line. A distribution unit is shared by spatially identical rows of all sub-macros. For example, row 0 of all $P$ sub-macros shares one distribution unit. The M value indicates the number of \textit{iActs} to be distributed by each multiplexer in the distribution unit. The N value of a sparsity pattern identifies the number of sub-macros that are aggregated together to process the same block M. In particular, for $N > 1$, multiple multiplexers will have the same set of inputs. Still, the distribution unit will select the appropriate input line based on the metadata (see \cref{sec:example} for a demonstration). The distribution unit abstracts the complexity of supporting large multiplexers in the memory cell by efficiently selecting the correct set of two activations to be fed to the memory array. We demonstrate in \cref{sec:pipeling} that by efficiently performing row and column pipelining, we require only 32 distribution units serving all rows of a column of all sub-macros every clock cycle.    

\noindent
\textbf{Merging unit. } The merging unit is responsible for accumulating the partial sums that are individually computed by each column of each sub-macro to generate the final partial sum in our partitioned scheme. To enable this, the merging unit comprises a $P$-input adder tree that reads its inputs from the partial sum accumulator's buffer.

\subsection{Pipelining Scheme}
\label{sec:pipeling}
Following on-chip buffers employed in comparable real-world systems \cite{farshchi2019integrating, sumbul2023fully}, we design the local \textit{iAct} buffer with a bandwidth of 1024 bits/cycle (128 8-bit activations). For dense LLM inference, 128 \textit{iAct}s are streamed to a column, enabling parallel MAC operations by all rows in a column of all sub-macros. However, for sparse inference with the highest sparsity \emph{i.e.}, 1:8 in our implementation, each row requires 8 \textit{iAct}s to choose from, which translates to 1024 \textit{iAct}s required by each column. Considering the upper limit of memory bandwidth, during 1:8 inference, only 4 rows of a column can perform parallel MAC operations. Similarly, 8 rows for 1:4 and 16 rows for 1:2. Due to the bit-serial nature of DCiMs, a MAC operation takes 8 cycles. Therefore, we overlap computation with memory access via row-pipelining. The number of row pipeline stages ($\#_{stages}$) is equal to 32/(\# of rows grouped). For example, for the 1:8 case, each set of 4 rows in a column--activated via \textit{EN\_COL}--are fed with \textit{iAct}s in a single cycle, and the remaining 28 rows are fed \textit{iAct}s in the subsequent 8 clock cycles in sets of 4, resulting in $\#_{stages}=8$. Every $\#_{stages}$ cycle, the \textit{EN\_COL} of the adjacent column is activated, and similarly, \textit{iAct}s are fed to all rows in the column. Once all columns have been fed with \textit{iAct}s, \textit{EN\_COL} of the first column is reactivated, and the process is repeated. We perform column-pipelining to reduce the overhead of the distribution unit because, without column pipelining, each column would require a dedicated distribution unit, which would be prohibitively expensive for memory-centric designs.     

\subsection{Walkthrough Example}
\label{sec:example}
To demonstrate FlexCiM enabling inference with diverse N:M patterns, we explain through two simple examples in \autoref{fig:flexcim_example}. The N of an N:M pattern identifies the number of sub-macros that work together to process a block of size M. Similarly, M specifies the number of \textit{iAct}s handled by each multiplexer in the distribution unit. For the 4:8 pattern, each set of four non-zero values are mapped to spatially identical locations in four sub-macros. For the 1:4 pattern, each sub-macro works independently on a block of size M, with each multiplexer receiving a separate set of 4 \textit{iAct}s. Note that the same distribution units employed for \textit{iAct}s can be used to map the weights to each of the sub-macros. 

\circled{1} The global controller receives the CSC metadata (in green) of the mapped weights and generates corresponding select signals to control the multiplexers. The LSB of the metadata is always reserved for the 2:1 multiplexer in the sub-macro. The remaining bits are employed for the multiplexers in the distribution unit. For example, for M=8, the metadata requires 3 bits, where the top 2 bits are for the distribution unit multiplexers. \circled{2} The distribution unit for each row (only row 0 unit shown in the figure) in column 0 selects the appropriate input line and directs it to the corresponding sub-macros. \circled{3} The first column of each of the sub-macros receives the selected \textit{iAct}s. These are then serialized and sent along the bit-lines. \circled{4} The 2:1 multiplexer selects the appropriate \textit{iAct} (highlighted in dark blue) from the bit-lines to be multiplied with the stored weights. \circled{5} After $\#_{stages}$ cycles, column 1 receives the \textit{iAct}s and the process is repeated. The merging unit performs final partial sum accumulation by reading from the PSum buffers of each column as they are calculated. Notably, in the 1:2 base case and dense inference, no selection occurs in the distribution units. Furthermore, for dense inference, the same \textit{iAct}s are streamed along both bit-lines, and $ I_{sel}$ is a don't-care.

\section{Results and Discussions}
\begingroup	
\begin{table}[t]\centering
 \caption{WikiText2 perplexity of pruning methods for different LLM model families at different target sparsity ratios.}
 \vspace{-5pt}
\resizebox{\linewidth}{!}
{%
\begin{tabular}{lc|ccc|cc|c|cc}
\Xhline{2\arrayrulewidth}
\rowcolor[HTML]{E0E0E0}
 &   & \multicolumn{3}{c|}{\textbf{LLaMA-2 \cite{touvron2023llama}}} &\multicolumn{2}{c|}{\textbf{LLaMA-3 \cite{meta2024introducing}}} & \multicolumn{1}{c|}{\textbf{Mixtral \cite{jiang2024mixtral}}} & \multicolumn{2}{c}{\textbf{Mamba \cite{gu2023mamba}}} \\
\cline{3-10}
\rowcolor[HTML]{E0E0E0}
 \raisebox{1.2ex}[1.2ex]{\textbf{Method}} & \raisebox{1.2ex}[1.2ex]{\textbf{N:M}} & \textbf{7B} & \textbf{13B} & \textbf{70B} & \textbf{8B} & \textbf{70B} & \textbf{8x7B} & \textbf{130M} & \textbf{1.4B} \\
\Xhline{2\arrayrulewidth}
\rowcolor[HTML]{D9EAF5}
Baseline & Dense & 5.47 & 4.83 & 3.31 & 6.13 & 2.85 & 3.84 & 20.04 & 10.42  \\
\Xhline{1\arrayrulewidth}
\rowcolor[HTML]{E0E0E0}
\multicolumn{10}{c}{Target Sparsity: $\sim 50\%$} \\
\Xhline{1\arrayrulewidth}
Magnitude & 4:8 & 16.87 & 9.59 & 6.83 & 9.16 & 5.84 & 7.91 & 58.63 & 43.24  \\
SparseGPT \cite{frantar2023sparsegpt} & 4:8 & 8.59 & 7.18 & 5.02 & 8.51 & 4.34 & 7.15 & 29.67 & 14.98 \\
Wanda \cite{sun2023simple} & 4:8 & 8.01 & 6.55 & 4.68 & 8.07 & 4.15 & 6.98 & 40.36 & 21.36 \\
OWL \cite{yin2023outlier} & Mixed N:8 & 7.92 & 6.08 & 4.55 & 8.02 & 3.95 & 6.57 & 27.28 & 14.07\\
\rowcolor[HTML]{D5E8D4}
FLOW (Ours) & Mixed N:M & \textbf{7.04} & \textbf{5.73} & \textbf{3.97} & \textbf{7.03} & \textbf{3.46} & \textbf{5.96} & \textbf{24.48} & \textbf{12.96}  \\
\Xhline{1\arrayrulewidth}
\rowcolor[HTML]{E0E0E0}
\multicolumn{10}{c}{Target Sparsity: $\sim 60\%$} \\
\Xhline{1\arrayrulewidth}
Magnitude & 3:8 & 54.59 & 48.96 & 49.51 & 52.38 & 57.63 & 56.76 & 1e20 & 1e14  \\
SparseGPT \cite{frantar2023sparsegpt} & 3:8 & 44.68 & 42.69 & 39.57 & 45.09 & 35.27 & 41.37 & 54.83  & 31.57  \\
Wanda \cite{sun2023simple} & 3:8 & 43.26 & 40.24 & 38.97 & 44.93 & 33.78 & 39.86 & 1268 & 1107 \\
OWL \cite{yin2023outlier} & Mixed N:8 & 20.68 & 18.68 & 18.57 & 16.26 & 11.23 & 14.58 & 31.56 & 18.67 \\
\rowcolor[HTML]{D5E8D4}
FLOW (Ours) & Mixed N:M & \textbf{18.96} & \textbf{16.53} & \textbf{17.95} & \textbf{15.24} & \textbf{10.89} & \textbf{13.83} & \textbf{28.61} & \textbf{15.37} \\
\Xhline{1\arrayrulewidth}
\rowcolor[HTML]{E0E0E0}
\multicolumn{10}{c}{Target Sparsity: Unconstrained} \\
\Xhline{1\arrayrulewidth}
OWL \cite{yin2023outlier} & Mixed N:8 (51.3\%) & 7.98 & 6.25 & 4.87 & 8.36 & 4.09 & 6.87 & 27.35 & 14.03 \\
\rowcolor[HTML]{D5E8D4}
FLOW (Ours) & Mixed N:M (58.5\%) & \textbf{7.13} & \textbf{5.95} & \textbf{4.57} & \textbf{8.03} & \textbf{3.87} & \textbf{6.35} & \textbf{25.12} & \textbf{13.08} \\
 \Xhline{2\arrayrulewidth}
\end{tabular}
}
\vspace{-6mm}
\label{tab:llm_results}
\end{table}
\endgroup
\subsection{Experimental Setup}
\noindent
\textbf{Models and datasets. }We evaluate transformer-based LLMs and VLMs \cite{meta2024introducing, jiang2024mixtral, touvron2023llama, awadalla2023openflamingo}, and also include Mamba-based SSMs \cite{gu2023mamba} for algorithm benchmarking. For pruning, we use 256 samples from PILE \cite{gao2020pile}. Models are compared by perplexity (PPL) on WikiText2 \cite{merity2018analysis} and accuracy on downstream tasks \cite{zellers2019hellaswag, clark2019boolq, chen2015microsoft, singh2019towards, bisk2020piqa}.

\noindent
\textbf{Algorithm implementation. }We implement FLOW in PyTorch. All experiments are conducted using a single NVIDIA H100 GPU. The runtime for the complete process of FLOW is less than 25 minutes, even for the largest model in our evaluation (LLaMA3 70B). 

\noindent
\textbf{Algorithm baselines. }We employ standard magnitude-based pruning as the naive baseline, and three SoTA pruning techniques, namely SparseGPT \cite{frantar2023sparsegpt}, Wanda \cite{sun2023simple}, and OWL \cite{yin2023outlier} to compare with FLOW. 

\noindent
\textbf{Accelerator implementation. }The FlexCiM accelerator is implemented in Verilog RTL and synthesized, placed-and-routed using Synopsys Design Compiler and Cadence Innovus, respectively, using TSMC 28nm technology library. In our experiments, we observed the RTL implementation of DCiM with optimized relative placement of logic cells and memory, results in PPA numbers similar to the case where logic is embedded within the memory cells (hardening of the memory macro). We chose the RTL for ease of experimentation, flexibility, and fair comparison with prior work. FlexCiM achieves a peak clock frequency of 1 GHz. Following \cite{sumbul2023fully, tu2022sdp}, we employ a two-level memory hierarchy: a level-1 SRAM that feeds data directly to the accelerator and a level-2 global shared SRAM that loads data from an off-chip DRAM. We assume all model parameters fit within the global SRAM for simplicity \cite{sumbul2023fully}. We use CACTI to estimate the area and power of level-1 and level-2 SRAMs. For end-to-end performance and energy metrics, we design a cycle-accurate simulator based on DNNWeaver \cite{sharma2016dnnweaver}, following prior works \cite{ramachandran2024algorithm}. 

\noindent
\textbf{Accelerator baselines. }We compare FlexCiM against SoTA sparse digital accelerator VEGETA \cite{jeong2023vegeta}, 1:2 structured sparse DCiM-based accelerator (SDP) and dense systolic array and dense DCiM. To ensure a fair comparison, we implement configurations of all accelerators so that they achieve iso-throughput. All digital accelerators are implemented with $64\times64$ array configuration to achieve the same peak throughput as DCiM-based designs with $128\times 32 \times 8$ configuration. All designs are scaled to 28nm using DeepScale \cite{sarangi2021deepscaletool}.

\subsection{FLOW: Results and Analysis}
\noindent
\textbf{Perplexity benchmark. } In \autoref{tab:llm_results}, we compare FLOW with the baselines for different target sparsity ratios. Magnitude, SparseGPT, and Wanda require static determination of the N:M pattern. We therefore assign 4:8 and 3:8 as the sparsity pattern for 50\% and 60\% target sparsity, respectively. Since OWL can explore different values of N for a fixed M, we fix M=8. Assigning M smaller than 8 for these methods results in poor performance. FLOW significantly outperforms the baselines, at all target sparsity ratios, \textbf{achieving up to $18\%$ perplexity improvement (PPL $\downarrow$)} over OWL, SparseGPT and Wanda. The benefit of FLOW is evident at higher target sparsity ratios, demonstrating that the higher sparse representational freedom offered by FLOW is instrumental in achieving better performance. 

Furthermore, we compare the performance of a FLOW-generated model with that generated via OWL for unconstrained target sparsity, where the framework has the freedom to choose the best N:M pattern for a layer that least impacts model performance. Notably, not only does FLOW result in a higher compression ($1.14times$), but it also achieves a lower PPL than OWL across all evaluated models. 

\begin{figure}[t]
    \centering \includegraphics[width=\columnwidth, keepaspectratio]{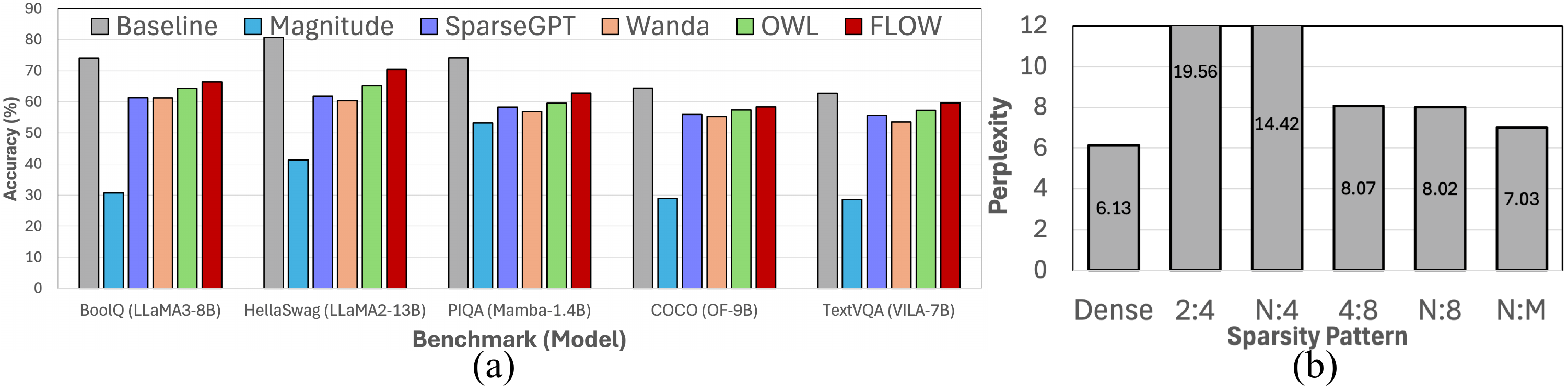}
        \vspace{-5mm}
        \caption{ (a) Accuracy (\%) comparison across 4 zero-shot tasks with 60\% target N:M sparsity across different LLMs and VLMs; (b) Ablation study on the impact of different N:M patterns.}
        \label{fig:llm_benchmark}
        \vspace{-5mm}
\end{figure}

\begin{figure}[t]
    \centering \includegraphics[width=0.8\columnwidth, keepaspectratio]{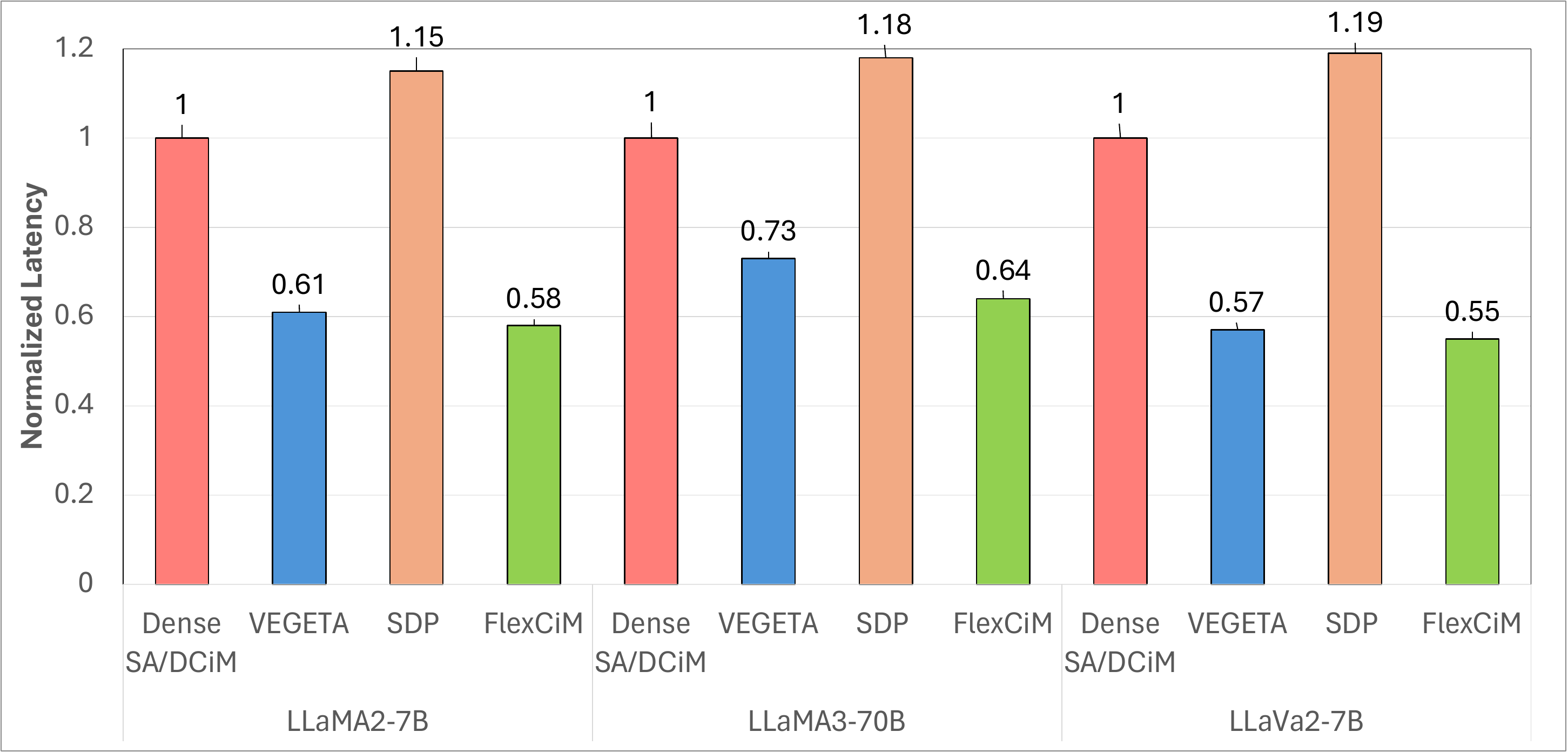}
        \vspace{-2mm}
        \caption{Normalized performance comparison between different dense and sparse accelerators for different models.}
        \label{fig:area}
        \vspace{-6mm}
\end{figure}

\noindent
\textbf{Zero-shot performance benchmark. }We evaluated the zero-shot performance of the pruned models in \autoref{fig:llm_benchmark}(a) at a target sparsity ratio of 60\%. In specific, we performed this evaluation with both LLMs and VLMs to study the generalization of FLOW across modalities. Notably, FLOW outperforms all other baselines across all tasks with an accuracy \textbf{improvement of up to 36\%}.

\noindent
\textbf{FLOW ablation. }We evaluate the impact of different N:M patterns on the perplexity with a LLaMA3-8B model in \autoref{fig:llm_benchmark}(b) for target sparsity of 50\%. To demonstrate the impact of static N:M choices, we performed ablations with two N:M patterns, namely 2:4 and 4:8, where the weights are pruned based on their importance scores. We find that the model pruned with 2:4 sparsity yields a significant increase in perplexity over the dense baseline. 4:8, though it improves the performance, still falls significantly behind the baseline dense. However, enabling flexibility for N, \emph{i.e.}, N:4 = \{1:4, 2:4, and 4:4 (dense)\} alleviates some of the perplexity degradation from a static N:M. More importantly, enabling flexibility for both N and M as proposed in FLOW \emph{i.e.}, N:M = \{1:2, 1:4, 2:4, 1:8, 2:8, 4:8, 8:8 (dense)\}, we achieve the lowest perplexity that is closest to the dense baseline. This shows the non-uniformity of LLM layers to sparsity that can be catered to only with a flexible selection of N and M.

\begingroup	
\setlength{\tabcolsep}{3 pt}
\begin{table}[t]\centering
 \caption{Comparison of FlexCiM with baselines at 28nm . \cmark,  \xmark~in a column identify the presence and absence of a feature, respectively.}
 \vspace{-6pt}
 \renewcommand*{\arraystretch}{1.0}
  \setlength\tabcolsep{1.9pt}
\resizebox{0.85\linewidth}{!}{%
\begin{tabular}{c|c|c|c|c|c}
\Xhline{2\arrayrulewidth}
 \textbf{Architecture} & \textbf{Component} (Area ($mm^2$)) & \makecell{\textbf{Total Area} \\ ($mm^2$)} & \makecell{\textbf{Flexible} \\ \textbf{N:M}} & \makecell{\textbf{Flexibility} \\ \textbf{Overhead}} & \makecell{\textbf{Compute} \textbf{Density} \\ (Peak TOPS/$mm^2$)} \\
  \Xhline{2\arrayrulewidth}

 \multirow{4}{*}{VEGETA \cite{jeong2023vegeta}} & Input/Output buffers (0.27)  & \multirow{4}{*}{3.28} & \multirow{4}{*}{\cmark} & \multirow{4}{*}{15.4\%} & \multirow{4}{*}{{2.36}} \\ \cline{2-2}
                                  & Weight buffer (0.18)& & & \\ \cline{2-2}
                                  & PE array (2.46)& & & \\ \cline{2-2}
                                  & Sparsity support (0.38)& & & \\ \cline{1-6}

 \Xhline{1\arrayrulewidth}

  \multirow{4}{*}{SDP \cite{tu2022sdp}} & Input/Output buffers (0.27)  & \multirow{4}{*}{0.98} & \multirow{4}{*}{\xmark} & \multirow{4}{*}{3.5\%} & \multirow{4}{*}{{7.65}} \\ \cline{2-2}
                                  & Weight buffer (0)& & & \\ \cline{2-2}
                                  & DCiM array (0.68)& & & \\ \cline{2-2}
                                  & Sparsity support (0.024)& & & \\ \cline{1-6}

 \Xhline{1\arrayrulewidth}

  \multirow{4}{*}{\textbf{FlexCiM} (Ours)}  & Input/Output buffers (0.27)  & \multirow{4}{*}{1.03} & \multirow{4}{*}{\cmark} & \multirow{4}{*}{5.9\%} & \multirow{4}{*}{7.28} \\ \cline{2-2}
                                  & Weight buffer (0)& & & \\ \cline{2-2}
                                  & DCiM array (0.72)& & & \\ \cline{2-2}
                                  & Sparsity support (0.043)& & & \\ \cline{1-6}
\Xhline{2\arrayrulewidth}
\end{tabular}
}
\label{table_accelerator_baseline}
\vspace{-7mm}
\end{table}
\endgroup
\subsection{FlexCiM: Results and Analysis}
\noindent
\textbf{Area comparison. }In \autoref{table_accelerator_baseline}, we compare the accelerator area breakdown of VEGETA and SDP with FlexCiM. All designs have the same buffer configurations. \textbf{SDP and FlexCiM have zero weight buffer area because these designs replace the weight memory and compute array of digital accelerators with DCiM macros}. FlexCiM and SDP have significantly lower area than VEGETA due to the high memory density of DCiM-based designs. In specific, FlexCiM, \textbf{despite supporting flexible N:M sparsity compared to SDP's fixed 1:2 sparsity, has a modest compute area overhead of 5.9\% and similar compute density}. Furthermore, FlexCiM's partitioned architecture reduces the long chain of column-wise adder trees in SDP by having multiple smaller adder trees across sub-macros.

\noindent
\textbf{Performance comparison. }In \autoref{fig:area}, we conduct a performance comparison across different models pruned via FLOW. We report normalized latency numbers, normalized to a dense systolic array (SA)/DCiM. Since we assume iso-peak-throughput accelerators, the dense SA and DCiM have same performance. Since SDP supports fixed 1:2 sparsity or dense, it has a higher latency than dense accelerators because FLOW assigns 1:2 pattern to very few layers, and in SDP, all other N:M patterns are treated as dense. \textbf{FlexCiM achieves a significantly lower latency up to 1.72$\times$  than the baseline accelerators.} This is due to the efficient acceleration of diverse N:M patterns and row-/column-pipelining scheme.

\noindent
\textbf{Energy comparison. }In \autoref{fig:energy_CiM}, we compare the normalized energy comparison of different accelerators running various models pruned via FLOW. We report energy breakdown in terms of local memory and global memory access, where local memory access measures energy from the Level-1 SRAM for DCiM designs and weight activation buffer for digital accelerators. The global memory access measures the energy consumption of data access from the shared SRAM. Since off-chip memory access will be approximately similar for DCiM or digital accelerators, we do not explicitly highlight it. Following \cite{sumbul2023fully}, we assume all model parameters fit in the global SRAM and consider DRAM access as out of scope for this work. FlexCiM has the lowest energy consumption across different models. Particularly, the DCiM-based architecture significantly reduces the energy contribution from local memory accesses. Furthermore, the bit-serial computation and efficient flexible N:M acceleration of FlexCiM also contribute to lower compute energy. \textbf{FlexCiM has up to 1.5$\times$ lower energy consumption compared to all baselines across different models.}

\section{Conclusion}
\label{sec:conc}
We introduced a novel N:M sparsity selection method dubbed as FLOW. We identify that outlier distribution in a layer contribute to their non-uniformity and tolerances to different sparsity ratios. FLOW enables the identification of optimal N and M values for each layer from a diverse set by accounting for both the presence and distribution of outliers. To accelerate such pruned models, we then proposed a DCiM-based accelerator,  FlexCiM. FlexCiM enables acceleration of diverse N:M patterns through a partitioned architecture, where the different DCiM macros are dynamically aggregated by a distribution and merging unit to support different N:M patterns.  Extensive experiments show the efficacy of FLOW over existing alternatives with an accuracy improvement of up to $36\%$, while FlexCiM delivers up to 1.75$\times$ lower inference latency.

\section*{Acknowledgments}
\noindent
This work was supported in part by CoCoSys, one of the seven centers in JUMP 2.0, a Semiconductor Research Corporation (SRC) program sponsored by DARPA. The authors would also like to thank the anonymous reviewers for their valuable feedback and suggestions, which helped improve the quality of this paper.

\bibliographystyle{IEEEtranS}
\footnotesize
\bibliography{main}

\begin{thebibliography}{10}
\providecommand{\url}[1]{#1}
\csname url@samestyle\endcsname
\providecommand{\newblock}{\relax}
\providecommand{\bibinfo}[2]{#2}
\providecommand{\BIBentrySTDinterwordspacing}{\spaceskip=0pt\relax}
\providecommand{\BIBentryALTinterwordstretchfactor}{4}
\providecommand{\BIBentryALTinterwordspacing}{\spaceskip=\fontdimen2\font plus
\BIBentryALTinterwordstretchfactor\fontdimen3\font minus \fontdimen4\font\relax}
\providecommand{\BIBforeignlanguage}[2]{{%
\expandafter\ifx\csname l@#1\endcsname\relax
\typeout{** WARNING: IEEEtranS.bst: No hyphenation pattern has been}%
\typeout{** loaded for the language `#1'. Using the pattern for}%
\typeout{** the default language instead.}%
\else
\language=\csname l@#1\endcsname
\fi
#2}}
\providecommand{\BIBdecl}{\relax}
\BIBdecl

\bibitem{agrawal2024metron}
A.~Agrawal, A.~Agarwal, N.~Kedia, J.~Mohan, S.~Kundu, N.~Kwatra, R.~Ramjee, and A.~Tumanov, ``Metron: Holistic performance evaluation framework for llm inference systems,'' \emph{arXiv preprint:2407.07000}, 2024.

\bibitem{awadalla2023openflamingo}
A.~Awadalla, I.~Gao, J.~Gardner, J.~Hessel, Y.~Hanafy, W.~Zhu, K.~Marathe, Y.~Bitton, S.~Gadre, S.~Sagawa \emph{et~al.}, ``Openflamingo: An open-source framework for training large autoregressive vision-language models,'' \emph{arXiv preprint arXiv:2308.01390}, 2023.

\bibitem{bisk2020piqa}
Y.~Bisk, R.~Zellers, J.~Gao, Y.~Choi \emph{et~al.}, ``Piqa: Reasoning about physical commonsense in natural language,'' in \emph{Proceedings of the AAAI conference on artificial intelligence}, vol.~34, no.~05, 2020, pp. 7432--7439.

\bibitem{chen2015microsoft}
X.~Chen, H.~Fang, T.-Y. Lin, R.~Vedantam, S.~Gupta, P.~Doll{\'a}r, and C.~L. Zitnick, ``Microsoft coco captions: Data collection and evaluation server,'' \emph{arXiv preprint arXiv:1504.00325}, 2015.

\bibitem{chen202115}
Z.~Chen, X.~Chen, and J.~Gu, ``15.3 a 65nm 3t dynamic analog ram-based computing-in-memory macro and cnn accelerator with retention enhancement, adaptive analog sparsity and 44tops/w system energy efficiency,'' in \emph{2021 IEEE International Solid-State Circuits Conference (ISSCC)}, vol.~64.\hskip 1em plus 0.5em minus 0.4em\relax IEEE, 2021, pp. 240--242.

\bibitem{clark2019boolq}
C.~Clark, K.~Lee, M.-W. Chang, T.~Kwiatkowski, M.~Collins, and K.~Toutanova, ``Boolq: Exploring the surprising difficulty of natural yes/no questions,'' \emph{arXiv preprint arXiv:1905.10044}, 2019.

\bibitem{duan2024towards}
C.~Duan \emph{et~al.}, ``Towards efficient sram-pim architecture design by exploiting unstructured bit-level sparsity,'' \emph{arXiv preprint arXiv:2404.09497}, 2024.

\bibitem{farshchi2019integrating}
F.~Farshchi, Q.~Huang, and H.~Yun, ``Integrating nvidia deep learning accelerator (nvdla) with risc-v soc on firesim,'' in \emph{2019 2nd Workshop on Energy Efficient Machine Learning and Cognitive Computing for Embedded Applications (EMC2)}.\hskip 1em plus 0.5em minus 0.4em\relax IEEE, 2019, pp. 21--25.

\bibitem{frantar2023sparsegpt}
E.~Frantar and D.~Alistarh, ``Sparse{GPT}: Massive language models can be accurately pruned in one-shot,'' 2023.

\bibitem{fujiwara20225}
H.~Fujiwara \emph{et~al.}, ``A 5-nm 254-tops/w 221-tops/mm 2 fully-digital computing-in-memory macro supporting wide-range dynamic-voltage-frequency scaling and simultaneous mac and write operations,'' in \emph{2022 IEEE International Solid-State Circuits Conference (ISSCC)}, vol.~65.\hskip 1em plus 0.5em minus 0.4em\relax IEEE, 2022, pp. 1--3.

\bibitem{gao2020pile}
L.~Gao, S.~Biderman, S.~Black, L.~Golding, T.~Hoppe, C.~Foster, J.~Phang, H.~He, A.~Thite, N.~Nabeshima \emph{et~al.}, ``The pile: An 800gb dataset of diverse text for language modeling,'' \emph{arXiv preprint arXiv:2101.00027}, 2020.

\bibitem{gu2023mamba}
A.~Gu and T.~Dao, ``Mamba: Linear-time sequence modeling with selective state spaces,'' \emph{arXiv preprint arXiv:2312.00752}, 2023.

\bibitem{jang2021sparsity}
J.-W. Jang \emph{et~al.}, ``Sparsity-aware and re-configurable npu architecture for samsung flagship mobile soc,'' in \emph{ISCA}, 2021, pp. 15--28.

\bibitem{jeong2023vegeta}
G.~Jeong, S.~Damani, A.~R. Bambhaniya, E.~Qin, C.~J. Hughes, S.~Subramoney, H.~Kim, and T.~Krishna, ``Vegeta: Vertically-integrated extensions for sparse/dense gemm tile acceleration on cpus,'' in \emph{2023 IEEE International Symposium on High-Performance Computer Architecture (HPCA)}.\hskip 1em plus 0.5em minus 0.4em\relax IEEE, 2023, pp. 259--272.

\bibitem{jiang2024mixtral}
A.~Q. Jiang, A.~Sablayrolles, A.~Roux, A.~Mensch, B.~Savary, C.~Bamford, D.~S. Chaplot, D.~d.~l. Casas, E.~B. Hanna, F.~Bressand \emph{et~al.}, ``Mixtral of experts,'' \emph{arXiv preprint arXiv:2401.04088}, 2024.

\bibitem{kim2021colonnade}
H.~Kim \emph{et~al.}, ``Colonnade: A reconfigurable sram-based digital bit-serial compute-in-memory macro for processing neural networks,'' \emph{IEEE Journal of Solid-State Circuits}, vol.~56, no.~7, pp. 2221--2233, 2021.

\bibitem{li2023laxor}
D.~Li, T.~Yamasaki, A.~Mani, A.~T. Do, N.~Chen, and B.~Wang, ``Laxor: A bit-accurate bnn accelerator with latch-xor logic for local computing,'' in \emph{2023 IEEE/ACM International Symposium on Low Power Electronics and Design (ISLPED)}.\hskip 1em plus 0.5em minus 0.4em\relax IEEE, 2023, pp. 1--6.

\bibitem{lin2024awq}
J.~Lin, J.~Tang, H.~Tang, S.~Yang, W.-M. Chen, W.-C. Wang, G.~Xiao, X.~Dang, C.~Gan, and S.~Han, ``Awq: Activation-aware weight quantization for on-device llm compression and acceleration,'' \emph{Proceedings of Machine Learning and Systems}, vol.~6, pp. 87--100, 2024.

\bibitem{liu202033}
Q.~Liu, B.~Gao, P.~Yao, D.~Wu, J.~Chen, Y.~Pang, W.~Zhang, Y.~Liao, C.-X. Xue, W.-H. Chen \emph{et~al.}, ``33.2 a fully integrated analog reram based 78.4 tops/w compute-in-memory chip with fully parallel mac computing,'' in \emph{2020 IEEE International Solid-State Circuits Conference-(ISSCC)}.\hskip 1em plus 0.5em minus 0.4em\relax IEEE, 2020, pp. 500--502.

\bibitem{liu2022s2ta}
Z.-G. Liu, P.~N. Whatmough, Y.~Zhu, and M.~Mattina, ``S2ta: Exploiting structured sparsity for energy-efficient mobile cnn acceleration,'' in \emph{2022 IEEE International Symposium on High-Performance Computer Architecture (HPCA)}.\hskip 1em plus 0.5em minus 0.4em\relax IEEE, 2022, pp. 573--586.

\bibitem{merity2018analysis}
S.~Merity, N.~S. Keskar, and R.~Socher, ``An analysis of neural language modeling at multiple scales,'' \emph{arXiv preprint arXiv:1803.08240}, 2018.

\bibitem{meta2024introducing}
A.~Meta, ``Introducing meta llama 3: The most capable openly available llm to date,'' \emph{Meta AI}, 2024.

\bibitem{mishra2021accelerating}
A.~Mishra, J.~A. Latorre, J.~Pool, D.~Stosic, D.~Stosic, G.~Venkatesh, C.~Yu, and P.~Micikevicius, ``Accelerating sparse deep neural networks,'' \emph{arXiv preprint arXiv:2104.08378}, 2021.

\bibitem{parashar2017scnn}
A.~Parashar, M.~Rhu, A.~Mukkara, A.~Puglielli, R.~Venkatesan, B.~Khailany, J.~Emer, S.~W. Keckler, and W.~J. Dally, ``Scnn: An accelerator for compressed-sparse convolutional neural networks,'' \emph{ACM SIGARCH computer architecture news}, vol.~45, no.~2, pp. 27--40, 2017.

\bibitem{raha2024flexnn}
A.~Raha, D.~A. Mathaikutty, S.~K. Ghosh, and S.~Kundu, ``Flex{NN}: A dataflow-aware flexible deep learning accelerator for energy-efficient edge devices,'' \emph{arXiv preprint:2403.09026}, 2024.

\bibitem{ramachandran2024microscopiq}
A.~Ramachandran, S.~Kundu, and T.~Krishna, ``Micro{S}copi{Q}: Accelerating foundational models through outlier-aware microscaling quantization,'' \emph{arXiv preprint arXiv:2411.05282}, 2024.

\bibitem{ramachandran2024algorithm}
A.~Ramachandran, Z.~Wan, G.~Jeong, J.~Gustafson, and T.~Krishna, ``Algorithm-hardware co-design of distribution-aware logarithmic-posit encodings for efficient dnn inference,'' \emph{arXiv:2403.05465}, 2024.

\bibitem{sarangi2021deepscaletool}
S.~Sarangi and B.~Baas, ``Deepscaletool: A tool for the accurate estimation of technology scaling in the deep-submicron era,'' in \emph{2021 IEEE ISCAS}.\hskip 1em plus 0.5em minus 0.4em\relax IEEE, 2021, pp. 1--5.

\bibitem{sharma2016dnnweaver}
H.~Sharma, J.~Park, E.~Amaro, B.~Thwaites, P.~Kotha, A.~Gupta, J.~K. Kim, A.~Mishra, and H.~Esmaeilzadeh, ``Dnnweaver: From high-level deep network models to fpga acceleration,'' in \emph{the Workshop on Cognitive Architectures}, 2016.

\bibitem{singh2019towards}
A.~Singh, V.~Natarajan, M.~Shah, Y.~Jiang, X.~Chen, D.~Batra, D.~Parikh, and M.~Rohrbach, ``Towards vqa models that can read,'' in \emph{Proceedings of the IEEE/CVF conference on computer vision and pattern recognition}, 2019, pp. 8317--8326.

\bibitem{sridharan2024sp}
A.~Sridharan, F.~Zhang, J.-S. Seo, and D.~Fan, ``Sp-imc: A sparsity aware in-memory-computing macro in 28nm cmos with configurable sparse representation for highly sparse dnn workloads,'' in \emph{2024 IEEE Custom Integrated Circuits Conference (CICC)}.\hskip 1em plus 0.5em minus 0.4em\relax IEEE, 2024, pp. 1--2.

\bibitem{sumbul2023fully}
H.~E. Sumbul, J.-s. Seo, D.~H. Morris, and E.~Beigne, ``A fully-digital and row-pipelined compute-in-memory neural network accelerator with soc-level benchmarking for ar/vr applications,'' \emph{IEEE Micro}, 2023.

\bibitem{sun2023simple}
M.~Sun, Z.~Liu, A.~Bair, and J.~Z. Kolter, ``A simple and effective pruning approach for large language models,'' \emph{ICLR}, 2024.

\bibitem{touvron2023llama}
H.~Touvron \emph{et~al.}, ``Llama 2: Open foundation and fine-tuned chat models,'' \emph{arXiv preprint:2307.09288}, 2023.

\bibitem{tu2022sdp}
F.~Tu, Y.~Wang, L.~Liang, Y.~Ding, L.~Liu, S.~Wei, S.~Yin, and Y.~Xie, ``Sdp: Co-designing algorithm, dataflow, and architecture for in-sram sparse nn acceleration,'' \emph{IEEE Transactions on Computer-Aided Design of Integrated Circuits and Systems}, vol.~42, no.~1, pp. 109--121, 2022.

\bibitem{wu2023highlight}
Y.~N. Wu, P.-A. Tsai, S.~Muralidharan, A.~Parashar, V.~Sze, and J.~Emer, ``Highlight: Efficient and flexible dnn acceleration with hierarchical structured sparsity,'' in \emph{Proceedings of the 56th Annual IEEE/ACM International Symposium on Microarchitecture}, 2023, pp. 1106--1120.

\bibitem{yin2023junk}
L.~Yin, S.~Liu, A.~Jaiswal, S.~Kundu, and Z.~Wang, ``A task-centric angle of llm pre-trained weights through sparsity,'' \emph{ICML}, 2024.

\bibitem{yin2023outlier}
L.~Yin, Y.~Wu, Z.~Zhang, C.-Y. Hsieh, Y.~Wang, Y.~Jia, G.~Li, A.~Jaiswal, M.~Pechenizkiy, Y.~Liang \emph{et~al.}, ``Outlier weighed layerwise sparsity: A missing secret sauce for pruning llms to high sparsity,'' \emph{ICML}, 2024.

\bibitem{yu2021compute}
S.~Yu, H.~Jiang, S.~Huang, X.~Peng, and A.~Lu, ``Compute-in-memory chips for deep learning: Recent trends and prospects,'' \emph{IEEE circuits and systems magazine}, vol.~21, no.~3, pp. 31--56, 2021.

\bibitem{yue202115}
J.~Yue \emph{et~al.}, ``15.2 a 2.75-to-75.9 tops/w computing-in-memory nn processor supporting set-associate block-wise zero skipping and ping-pong cim with simultaneous computation and weight updating,'' in \emph{2021 IEEE International Solid-State Circuits Conference (ISSCC)}, vol.~64.\hskip 1em plus 0.5em minus 0.4em\relax IEEE, 2021, pp. 238--240.

\bibitem{zellers2019hellaswag}
R.~Zellers, A.~Holtzman, Y.~Bisk, A.~Farhadi, and Y.~Choi, ``Hellaswag: Can a machine really finish your sentence?'' \emph{arXiv preprint arXiv:1905.07830}, 2019.

\bibitem{zhan202428}
Y.~Zhan, W.-H. Yu, K.-F. Un, R.~P. Martins, and P.-I. Mak, ``A 28-nm 18.7 tops/mm2 89.4-to-234.6 tops/w 8b single-finger edram compute-in-memory macro with bit-wise sparsity aware and kernel-wise weight update/refresh,'' \emph{IEEE Journal of Solid-State Circuits}, 2024.

\bibitem{zhang2023dynamic}
Y.~Zhang, L.~Zhao, M.~Lin, Y.~Sun, Y.~Yao, X.~Han, J.~Tanner, S.~Liu, and R.~Ji, ``Dynamic sparse no training: Training-free fine-tuning for sparse llms,'' \emph{arXiv preprint arXiv:2310.08915}, 2023.

\bibitem{zhong2023digital}
B.~Zhong, M.~Wang, C.~Zhang, Y.~Mai, X.~Li, and Z.~Yu, ``A digital sram computing-in-memory design utilizing activation unstructured sparsity for high-efficient dnn inference,'' in \emph{2023 IEEE Computer Society Annual Symposium on VLSI (ISVLSI)}.\hskip 1em plus 0.5em minus 0.4em\relax IEEE, 2023, pp. 1--6.

\end{thebibliography}

\end{document}